\documentclass[letterpaper, 10 pt, conference]{IEEEtran}
\IEEEoverridecommandlockouts    
\usepackage{hyperref}
\usepackage{threeparttable}
\usepackage{algorithm}
\usepackage{cite}
\usepackage{multirow}
\usepackage{booktabs}
\usepackage{xcolor}      
\usepackage{colortbl}    
\usepackage{pifont}      
\usepackage{tikz}
\usetikzlibrary{shapes, arrows, positioning, fit, calc, shadows}
\newcommand{\etal}{\textit{et al.}}
\hypersetup{
    colorlinks=true,
    linkcolor=magenta,
    filecolor=magenta,      
    urlcolor=magenta,
}
\hypersetup{
    colorlinks=true,
    linkcolor=magenta,
    filecolor=magenta,      
    urlcolor=magenta,
}

\usepackage{graphics}           
\usepackage{times}              
\usepackage{amsmath}            
\usepackage{amssymb}            
\usepackage{graphicx}
\usepackage{algorithm}
\usepackage[noend]{algpseudocode}
\usepackage{booktabs}
\usepackage{color}
\definecolor{instructioncolor}{rgb}{.5,.5,.5}

\usepackage[font=small]{caption}

\def\eqref#1{Eq.~(\ref{#1})}

\makeatletter
\usepackage{xspace}
\DeclareRobustCommand\onedot{\futurelet\@let@token\@onedot}
\def\@onedot{\ifx\@let@token.\else.\null\fi\xspace}

\def\etal{{et al}\onedot}
\makeatother

\usepackage{array}
\newcolumntype{L}[1]{>{\raggedright\let\newline\\\arraybackslash\hspace{0pt}}m{#1}}
\newcolumntype{C}[1]{>{\centering\let\newline\\\arraybackslash\hspace{0pt}}m{#1}}
\newcolumntype{R}[1]{>{\raggedleft\let\newline\\\arraybackslash\hspace{0pt}}m{#1}}

\usepackage{bm}
\usepackage{multirow}
\usepackage{rotating}  

\title{\LARGE \bf VGGT-MPR: VGGT-Enhanced Multimodal Place Recognition\\ in Autonomous Driving Environments}

\author{Jingyi Xu$^{1}$, Zhangshuo Qi$^{2}$, Zhongmiao Yan$^{1}$, Xuyu Gao$^{1}$, Qianyun Jiao$^{1}$,\\ Songpengcheng Xia$^{1}$, Xieyuanli Chen$^{3}$, and Ling Pei$^{1 *}$ 
  \thanks{This work was supported in part by the National Nature Science Foundation of China (NSFC)under Grant No.62273229,in part by the Science and Technology Commission of Shanghai Municipality under Grant No.24DZ3101300, No.24TS1402600, and No.24TS1402800 separately.}
  \thanks{$^{1}$Jingyi Xu, Zhongmiao Yan, Xuyu Gao, Qianyun Jiao, Songpengcheng Xia, and Ling Pei are with the Shanghai Jiao Tong University.}
  \thanks{$^{2}$Zhangshuo Qi is with the Beijing Institute of Technology.}
  \thanks{$^{3}$Xieyuanli Chen is with the National University of Defense Technology.}
  \thanks{$^*$Corresponding author: Ling Pei (ling.pei@sjtu.edu.cn)}
}

\begin{document}
\maketitle

\IEEEpeerreviewmaketitle
\thispagestyle{empty}
\pagestyle{empty}

\begin{abstract}
In autonomous driving, robust place recognition is critical for global localization and loop closure detection. While inter-modality fusion of camera and LiDAR data in multimodal place recognition (MPR) has shown promise in overcoming the limitations of unimodal counterparts, existing MPR methods basically attend to hand-crafted fusion strategies and heavily parameterized backbones that require costly retraining. To address this, we propose VGGT-MPR, a multimodal place recognition framework that adopts the Visual Geometry Grounded Transformer (VGGT) as a unified geometric engine for both global retrieval and re-ranking.
In the global retrieval stage, VGGT extracts geometrically-rich visual embeddings through prior depth-aware and point map supervision, and densifies sparse LiDAR point clouds with predicted depth maps to improve structural representation. This enhances the discriminative ability of fused multimodal features and produces global descriptors for fast retrieval.
Beyond global retrieval, we design a training-free re-ranking mechanism that exploits VGGT's cross-view keypoint-tracking capability. By combining mask-guided keypoint extraction with confidence-aware correspondence scoring, our proposed re-ranking mechanism effectively refines retrieval results without additional parameter optimization.
Extensive experiments on large-scale autonomous driving benchmarks and our self-collected data demonstrate that VGGT-MPR achieves state-of-the-art performance, exhibiting strong robustness to severe environmental changes, viewpoint shifts, and occlusions. 
Our code and data will be made publicly available.
\end{abstract}


\section{Introduction}
\label{sec:intro}
\IEEEPARstart{I}{n} autonomous driving applications, place recognition plays a crucial role in various downstream tasks like loop closure detection in simultaneous localization and mapping (SLAM) and global localization in GPS-denied environments~\cite{joseph2025matched, luo2025bevplace++, chen2021overlapnet}. 
Visual place recognition (VPR) relies on camera image inputs to retrieve similar locations from a reference database~\cite{keetha2023anyloc, wang2022salient, nie2024mixvpr++}. As cameras are cost-effective and widely available, VPR offers a practical and straightforward solution. However, visual data are susceptible to environmental variations such as illumination and weather, leading to performance instability.
LiDAR place recognition (LPR)~\cite{ma2022overlaptransformer, luo2025overlapmamba, ma2023cvtnet} can mitigate such issues but suffers from a lack of texture information, making it prone to noise and reduced precision.

To address these limitations, multimodal place recognition (MPR) methods have been developed to combine complementary strengths across modalities, emerging as a more robust approach for autonomous driving applications~\cite{lu2020pic, pan2021coral, komorowski2021minkloc++}.
However, the existing MPR methods have largely focused on hand-crafted fusion schemes and densely parameterized networks. This increases the difficulty of algorithm design and significantly reduces the deployment efficiency. Recently, a few VPR approaches~\cite{wang2022salient, nie2024mixvpr++, lu2024cricavpr} have attended to the out-of-the-box foundation models~\cite{radford2021learning, oquab2023dinov2, kirillov2023segment} that have a strong capability of visual feature extraction from camera images. They basically present impressive recognition performance to identify similar places with the single vision modality. However, how foundation models can be integrated into the MPR scheme to concurrently benefit coordinated modalities for higher recognition accuracy still remains unexplored.

\begin{figure}
  \centering
  \includegraphics[width=0.98\linewidth]{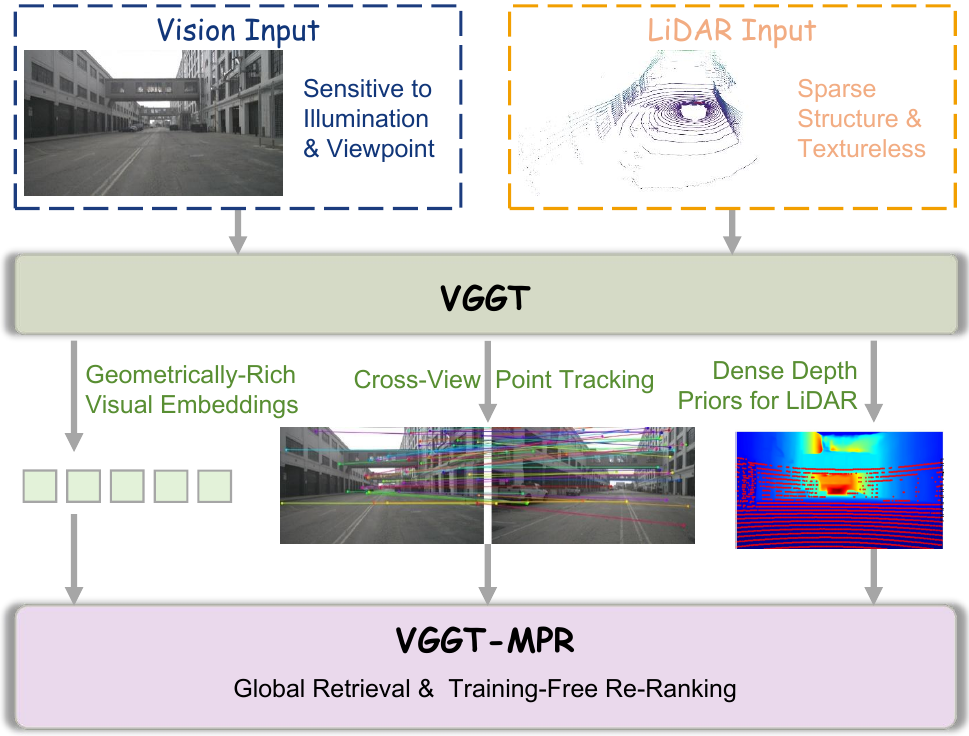}
  \caption{From VGGT to VGGT-MPR. VGGT is reinterpreted as a unified geometry-centric foundation to address modality-specific limitations in multimodal place recognition. It enhances visual representations with implicit structural awareness, densifies sparse LiDAR observations via depth estimation for global retrieval, and provides reliable cross-view point tracking for training-free re-ranking.}
  \label{fig:motivation}
  \vspace{-0.6cm}
\end{figure}

To bridge this gap, we present a novel perspective on reinterpreting the recently emerging foundation model, Visual Geometry Grounded Transformer (VGGT)~\cite{wang2025vggt}, as a unified geometric engine for multimodal place recognition. 
VGGT utilizes a feed-forward neural network capable of jointly inferring key 3D attributes of a scene from one or multiple views. It achieves state-of-the-art (SOTA) performance in multiple downstream tasks, including camera pose inference, depth estimation, point map generation, and image matching. In this work, instead of treating vision and LiDAR as separate streams fused only at the end, as Fig.~\ref{fig:motivation} highlights, we introduce VGGT as a geometry-centric foundation that concurrently upgrades both modalities. Our key insights are as follows: 
\begin{itemize}
\item \textbf{Geometrically-rich features for the vision branch.} Conventional vision branches in MPR exploit standard CNNs or transformers as parameterized backbones, which are trained from scratch and basically overlook essential geometric structures in driving scenes, such as building layouts, spatial configurations, and occlusion patterns. In contrast, VGGT is supervised concurrently by 3D structural signals, which can naturally provide geometrically-rich visual embeddings that enhance the discriminativeness of place description.

\item \textbf{Dense ranges for the LiDAR branch.} Existing MPR methods only encode sparse point clouds in the LiDAR branch, leading to limited structural awareness. VGGT generates dense point maps within the field of view, thus providing range information to densify sparse point clouds from LiDAR observations. This helps to enhance the MPR model's perception of spatial structural details.

\item \textbf{Cross-view consistency.} The cross-view consistency inherent in VGGT can benefit place retrieval by keypoint tracking. The alternating-attention mechanism of VGGT strategically alternates between frame-wise and global self-attentions, enabling robust intra-frame representations and subsequent interaction across all input views. Thus, VGGT achieves plausible point tracking across observations around the same place, which provides the re-ranking criterion to refine recognition results.

\item \textbf{Robustness to viewpoints and occlusions.} VGGT's training paradigm and network design inherently equip it with exceptional robustness to severe viewpoint changes and dynamic occlusions, which are critical challenges in real-world place recognition for autonomous driving.

\end{itemize}

In sum, the main contributions of our work are threefold:

\begin{itemize}

\item We propose VGGT-MPR, a novel framework that reinterprets VGGT as a unified geometric engine. To the best of our knowledge, this is the first work to leverage this visual foundation model to bridge visual perception, 3D environmental structures, and cross-view consistency for robust multimodal place recognition.

\item We introduce geometry-centric feature extraction where VGGT serves a dual purpose, i.e., extracting geometrically-rich visual embeddings and densifying sparse LiDAR point clouds via dense depth priors. This effectively enhances the discriminative ability of the global descriptors by capturing complementary strengths across modalities for fast retrieval.

\item We design a training-free re-ranking mechanism that exploits VGGT's robust cross-view point-tracking capability. By incorporating mask-guided keypoint extraction and confidence-aware correspondence scoring, this module refines retrieval results without requiring additional parameter optimization.

\end{itemize}


\section{Related Work}
\label{sec:related}

\subsection{Multimodal Place Recognition}
\label{sec: related_MPR}
MPR leverages complementary data from multiple sensors like cameras and LiDAR to achieve superior performance compared to unimodal counterparts. 
For example, MinkLoc++~\cite{komorowski2021minkloc++} learns to emphasize more reliable modality features during training, capturing modality dominance. To improve the balancing of modality contributions, AdaFusion~\cite{lai2022adafusion} generates adaptive weights via an attention mechanism to derive weighted fusions of image and point cloud descriptors. To further enhance yaw-invariance, LCPR~\cite{zhou2023lcpr} creates representations invariant to yaw rotation to increase robustness following OT series~\cite{chen2021overlapnet, ma2022overlaptransformer, ma2023cvtnet}. Differing from implicit fusion schemes, EINet~\cite{xu2024explicit} establishes explicit cross-modal interaction by depth supervision and color rendering. Most recently, GSPR~\cite{qi2024gspr} effectively integrates different modalities into a unified 3D-GS scene representation, but sacrifices real-time performance.
While these methods have advanced multimodal fusion, they largely overlook the off-the-shelf feature extraction capabilities of foundation models. Training the entire backbone from scratch is always required, leading to lower deployment efficiency in real applications. In contrast, in this work, we adopt the recently emerging foundation model as a unified geometric engine, which facilitates effective visual embedding extraction and point cloud densification, and enables training-free re-ranking refinement.

\begin{figure*}
  \centering
  \includegraphics[width=0.98\linewidth]{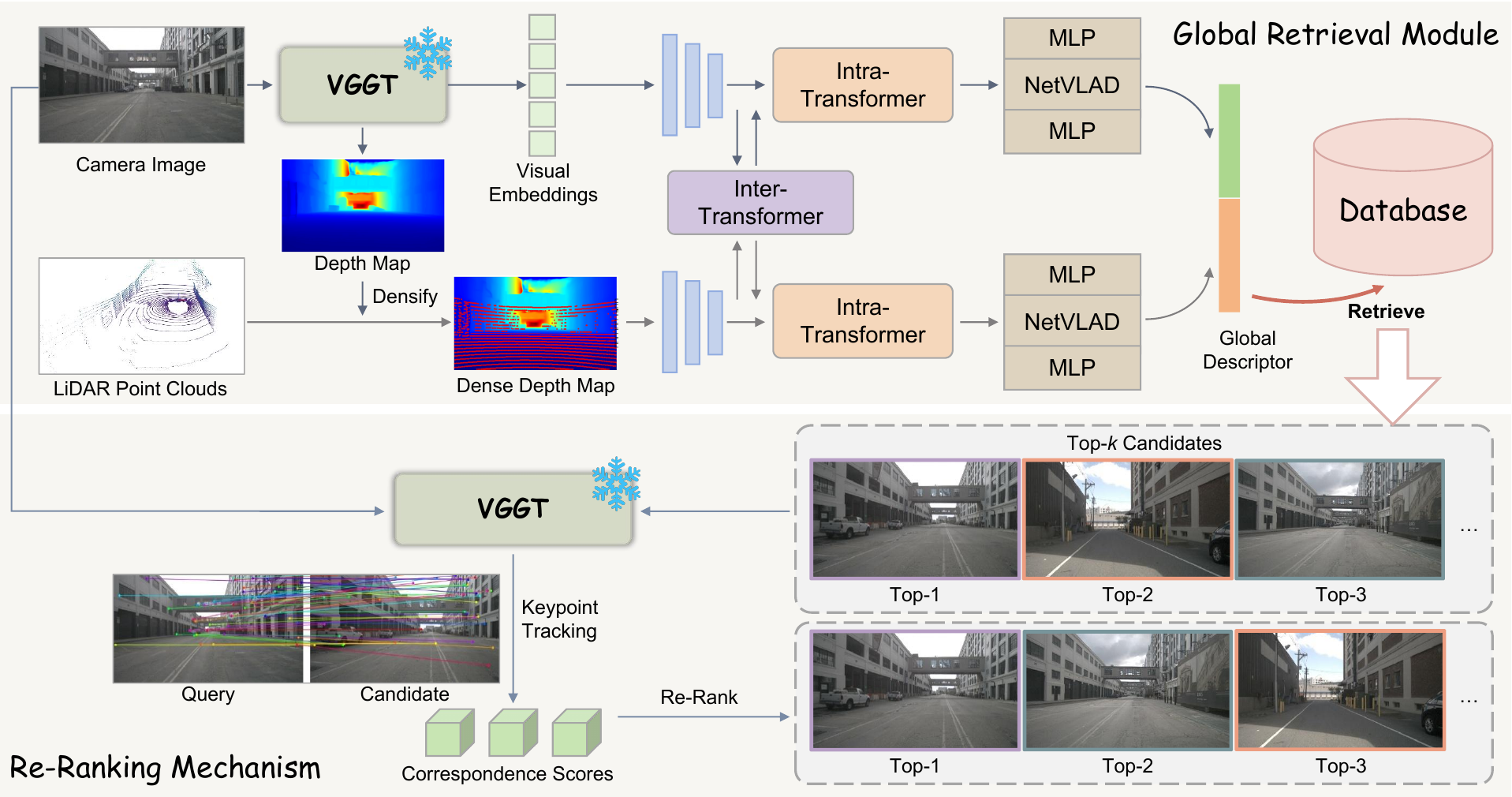}
  \caption{Overview of our proposed VGGT-MPR. Capitalizing on the spatial perception capabilities of VGGT, the pipeline integrates a global retrieval module (GRM) and a re-ranking mechanism (RRM). The GRM fuses multimodal inputs (camera images and LiDAR point clouds) to generate global descriptors for database indexing and retrieval. Subsequently, the RRM refines the retrieved candidates to improve place recognition accuracy.}
  \label{fig:pipeline}
  \vspace{-0.5cm}
\end{figure*}

\subsection{Foundation Model-Based Place Recognition}
\label{sec: related_FMPR}

Foundation models~\cite{radford2021learning, oquab2023dinov2, kirillov2023segment}, pretrained on large-scale diverse datasets, provide powerful and generalizable feature representations that can be effectively transferred to place recognition.
For example, Text2Loc~\cite{xia2024text2loc} and UniLoc~\cite{xia2024uniloc} extract textural features with the CLIP~\cite{radford2021learning} architecture to reduce the feature distance between language descriptions and semantic information.
In addition, SalientVPR~\cite{wang2022salient} leverages attention maps of DINOv2~\cite{oquab2023dinov2} to identify critical regions for retrieval. Similarly, LLM-VPR~\cite{lyu2024tell} leverages DINOv2 as a vision-based coarse retriever and further improves its performance with GPT-4V~\cite{achiam2023gpt}. Several recent methods including MixVPR$++$~\cite{nie2024mixvpr++}, SALAD~\cite{izquierdo2024optimal}, and CricaVPR~\cite{lu2024cricavpr} also build upon the DINOv2 backbone for visual feature extraction. In contrast, AnyLoc~\cite{keetha2023anyloc} explores the effectiveness of CLIP, DINOv2, and MAE~\cite{he2022masked} to capture long-range global patterns.
Compared to the above-mentioned foundation models, SAM~\cite{kirillov2023segment} offers another valuable capability by generating high-quality semantic masks, enabling place recognition systems to remove dynamic distractions. For example, SegVLAD~\cite{garg2024revisit} combines DINOv2 with SAM to generate SuperSegments along with segment-level descriptors. To further improve inference efficiency, Jiao~\etal~\cite{jiao2025inscmpr} use MobileSAM~\cite{zhang2023faster} to obtain cross-modal instance priors and suppress background noise. 
Despite these advances in foundation model-based approaches, the potential of VGGT remains largely unexplored in MPR. Given its strong spatial perception capability, VGGT offers a promising foundation for developing robust multimodal place recognition systems, which will be explored in this work.

\subsection{Re-Ranking for Place Recognition}
\label{sec: related_RFPR}

Re-ranking initially retrieved candidates is an effective strategy for improving place recognition performance. 
Some VPR methods, such as ETR~\cite{zhang2023etr}, perform re-ranking through capturing image-pair similarities with self- and cross-attention mechanisms, leveraging both global and local descriptors. Similarly, $R^2$Former~\cite{zhu2023r2former} uses attention maps from the transformer-based global retrieval module as local features for re-ranking.
In LPR, SpectralGV~\cite{vidanapathirana2023spectral} re-ranks candidates by evaluating the spatial consistency of local features. Inspired by this work, LPR-Mate~\cite{zhang2025lpr} triggers a graph-based ranking when spatial consistency match scores are lower than a preset threshold.
In contrast to the above methods focusing on local or hybrid features, TReR~\cite{barros2023trer} relies solely on global descriptors and reorders candidates via a transformer-based re-ranker.
Although these diverse re-ranking mechanisms have been shown to enhance recognition performance, a common limitation is their reliance on supervised training, which requires additional module optimization. Furthermore, all existing re-ranking mechanisms are designed for unimodal place recognition, and no prior work has integrated re-ranking into multimodal counterparts.
To address these gaps, we propose a novel re-ranking mechanism that leverages VGGT's strong point-tracking capability, eliminating the need for additional module optimization and refining the final recognition results.


\section{Methodology}
\label{sec:method}

\subsection{VGGT-MPR Architecture}
\label{sec:architecture}

As illustrated in Fig.~\ref{fig:pipeline}, our proposed VGGT-MPR exploits VGGT's robust spatial perception capabilities and comprises two main components: a global retrieval module (GRM) and a re-ranking mechanism (RRM). 
The GRM processes multimodal inputs, specifically the front-view camera image $\mathcal{I}$ and the LiDAR point clouds $\mathcal{P}$, to generate a unified global descriptor $\overline{\mathcal{D}}$. 
Once the network is trained, these generated descriptors are indexed to construct a reference database $\mathcal{M}$. 
During the inference phase, the generated global descriptor of a query frame serves as a probe to retrieve candidate matches $\mathcal{C}$ from $\mathcal{M}$. These candidates are subsequently processed by the RRM to refine the retrieval ranking, thereby further enhancing the place recognition performance.
In this pipeline, VGGT-MPR uses VGGT as a frozen geometry-centric backbone to enhance multimodal global retrieval, while reserving its tracking capability for an effective re-ranking stage.

\subsection{Global Retrieval Module}
\label{sec:GRM}

The GRM is designed to generate discriminative global descriptors by effectively fusing features from both camera and LiDAR modalities. In this module, the frozen VGGT backbone performs two tasks on raw input observations: visual embedding extraction and point cloud densification. 

Specifically, the input front-view camera image $\mathcal{I}\in\mathbb{R}^{3\times H\times W}$ is processed by VGGT to yield high-level visual embeddings $\mathcal{F}^v\in\mathbb{R}^{C\times H'\times W'}$ and a virtual depth map $\mathcal{T}^v\in\mathbb{R}^{1\times H\times W}$. 
Consistent with~\cite{wang2025vggt}, $\mathcal{F}^v$ consists of the output tokens from the VGGT, while $\mathcal{T}^v$ is predicted by the DPT head. As VGGT is pretrained through prior depth-aware and point map supervision to capture reasonable geometric features with effective attention combinations, $\mathcal{F}^v$ naturally encompasses structural information for spatial understanding. This facilitates the improvement of the discriminative visual embeddings for place description.
In addition, although $\mathcal{T}^v$ provides dense pixel-wise depth, it lacks a real-world metric scale. In contrast, the input LiDAR point clouds $\mathcal{P}\in\mathbb{R}^{3\times N_\text{pc}}$ capture accurate metric information but are inherently sparse. To address this complementarity, we utilize $\mathcal{T}^v$ to densify the sparse LiDAR data, generating a dense depth map $\mathcal{T}^s\in\mathbb{R}^{1\times H\times W}$ that incorporates real-world scale.
Specifically, we exploit the anchor-based scaling method, which selects the LiDAR points projected to the image plane as anchors. For each anchor with an absolute depth, the relative depth value at the corresponding image location is retrieved from the depth map predicted by VGGT. Using the matched pairs of relative and absolute depths, a scale relationship is estimated through linear regression across all laser points. Thus, we obtain the estimated scale and offset parameters, which are finally applied to the entire relative depth map, producing an absolute depth map $\mathcal{T}^s$ aligned with real-world metric depth values. As shown, VGGT serves as a geometry-centric backbone for processing different input modalities to generate the following descriptors.

Subsequently, the visual embeddings $\mathcal{F}^v$ and the metric dense depth map $\mathcal{T}^s$ are processed by lightweight convolution networks to extract intermediate features $\mathcal{F}^i$ and $\mathcal{F}^l$, respectively. 
To facilitate feature interaction between the two modalities, these features are passed through inter-transformer, yielding $\mathcal{F}^{i'}$ and $\mathcal{F}^{l'}$. Specifically, in the inter-transformer of the vision branch, we first generate $\mathcal{F}^{i'}$ with the query and key from $\mathcal{F}^i$ and value from $\mathcal{F}^l$. Concurrently, $\mathcal{F}^{l'}$ is produced by the inter-transformer of the LiDAR branch, which uses the query and key from $\mathcal{F}^l$ and value from $\mathcal{F}^i$.
Finally, we apply intra-transformer layers in both branches, which are self-attention of transformer, followed by NetVLAD with MLPs to aggregate the features into an image descriptor $\mathcal{D}^i\in\mathbb{R}^{D}$ and a LiDAR descriptor $\mathcal{D}^l\in\mathbb{R}^{D}$. They are concatenated to form the final global descriptor $\overline{\mathcal{D}}\in\mathbb{R}^{2D}$.

\begin{table*}[t]
\scriptsize
\setlength{\tabcolsep}{6.3pt}
\center
\captionsetup{aboveskip=2pt, belowskip=0pt}
\renewcommand\arraystretch{0.95}
\caption{Performance comparison on the nuScenes dataset}
\begin{tabular}{l|c|cccc|cccc|cccc}
\toprule
\multicolumn{1}{l|}{\multirow{2}{*}{Approach}}  & \multicolumn{1}{l|}{\multirow{2}{*}{Modality}}   & \multicolumn{4}{c|}{BS Split} & \multicolumn{4}{c|}{SON Split}   & \multicolumn{4}{c}{SQ Split}  \\ \cmidrule{3-14} 
\multicolumn{1}{c|}{} & \multicolumn{1}{c|}{} & AR@1  & AR@5  & AR@10  & AR@20  & AR@1  & AR@5  & AR@10  & AR@20  & AR@1  & AR@5  & AR@10  & AR@20\\ \cmidrule{1-14}
PatchNetVLAD & V  & 83.41 & 89.96 & 94.21 & 96.75 & 95.34 & 97.97 & 98.82 & 90.01 & 86.92 & 88.94 & 89.42 & 91.30 \\
CricaVPR & V  & 87.10 & 97.34 & 99.30 & \underline{99.86} & 93.94 & 98.74 & 98.99 & \underline{99.75} & 87.80 & 91.46 & 92.68 & 94.21 \\
CVTNet   & L  & 86.09 & 94.96 & 96.88 & 97.36 & 98.04 & 98.16 & 99.45 & 99.45 & \underline{98.93} & \underline{99.40}  & \underline{99.75} & \textbf{100.00} \\
OverlapMamba  & L  & 74.58 & 88.28 & 93.29 & 96.64 & 86.87 & 97.47 & 98.48 & 98.74 & 95.73 & 99.39 & 99.70 & \textbf{100.00} \\
\cmidrule{1-14}
MinkLoc++  & V+L  & 76.72 & 92.57 & 96.07 & 97.19 & 89.39 & 98.74 & 99.49 & \textbf{100.00} & 89.94 & 96.04 & 97.87 & 98.48 \\
LCPR   & V+L  & 84.65 & 96.16 & 97.12 & 98.56 & 92.42 & 97.98 & 99.24 & 99.49 & 87.20 & 94.82 & 95.12 & 96.04 \\
EINet  & V+L  & \underline{91.40} & 97.30 & 98.75 & 99.39 & \underline{98.29} & \underline{98.91} & \underline{99.66} & 99.70 & 73.84 & 85.50 & 91.37 & 93.31 \\
GSPR   & V+L  & 90.32 & \underline{98.88} & \underline{99.58} & 99.58 & 88.89 & 98.74 & 99.49 & \underline{99.75} & 90.55 & 98.78 & 98.78 & \underline{99.39} \\
\rowcolor{lightgray} VGGT-MPR (ours)  & V+L  & \textbf{98.28} & \textbf{99.76} & \textbf{99.93} & \textbf{99.93} & \textbf{99.74} & \textbf{100.00} & \textbf{100.00} & \textbf{100.00} & \textbf{99.28} & \textbf{99.80} & \textbf{99.80} & \textbf{100.00} \\ \bottomrule
\end{tabular}
\label{tab: comparison}
\\ \vspace{-0.15cm}
\begin{flushleft}
\scriptsize
V: Vision, L: LiDAR, V+L: Vision+LiDAR. The best and secondary results are highlighted in \textbf{bold black} and \underline{underline}, respectively.
\end{flushleft}
\vspace{-0.7cm}
\end{table*}

\begin{figure}[t]
  \centering
  \includegraphics[width=0.98\linewidth]{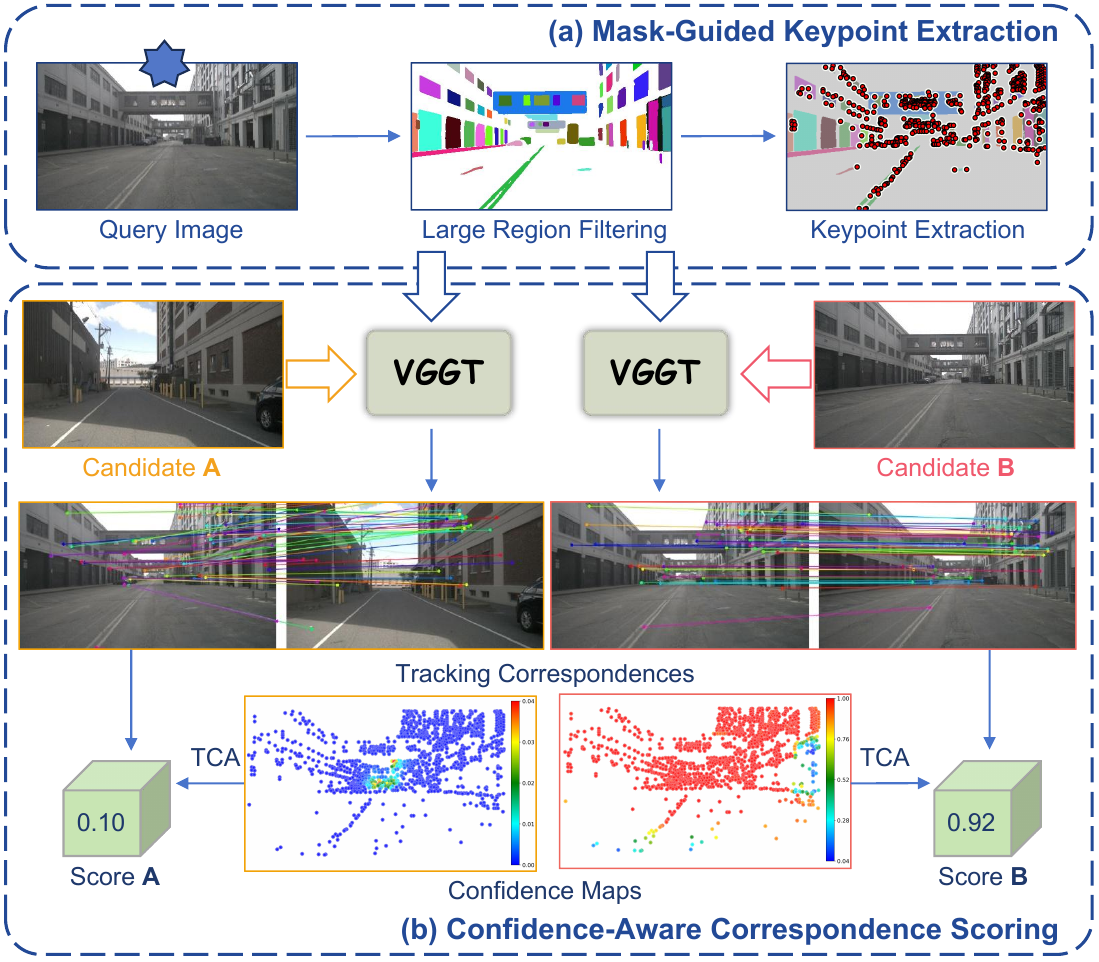}
  \caption{
  Our proposed re-ranking mechanism in VGGT-MPR. Given a query image and a corresponding candidate from the top-$k$ matches (e.g., candidates \textbf{A} or \textbf{B}), we perform mask-guided keypoint extraction (a) and confidence-aware correspondence scoring (b) to calculate the score for the input candidate. The top-$k$ candidates are ultimately re-ranked based on their respective correspondence scores to enhance recognition accuracy.
  }
  \label{fig:re-ranking}
  \vspace{-0.4cm}
\end{figure}

\begin{figure}[t]
  \centering
  \includegraphics[width=0.98\linewidth]{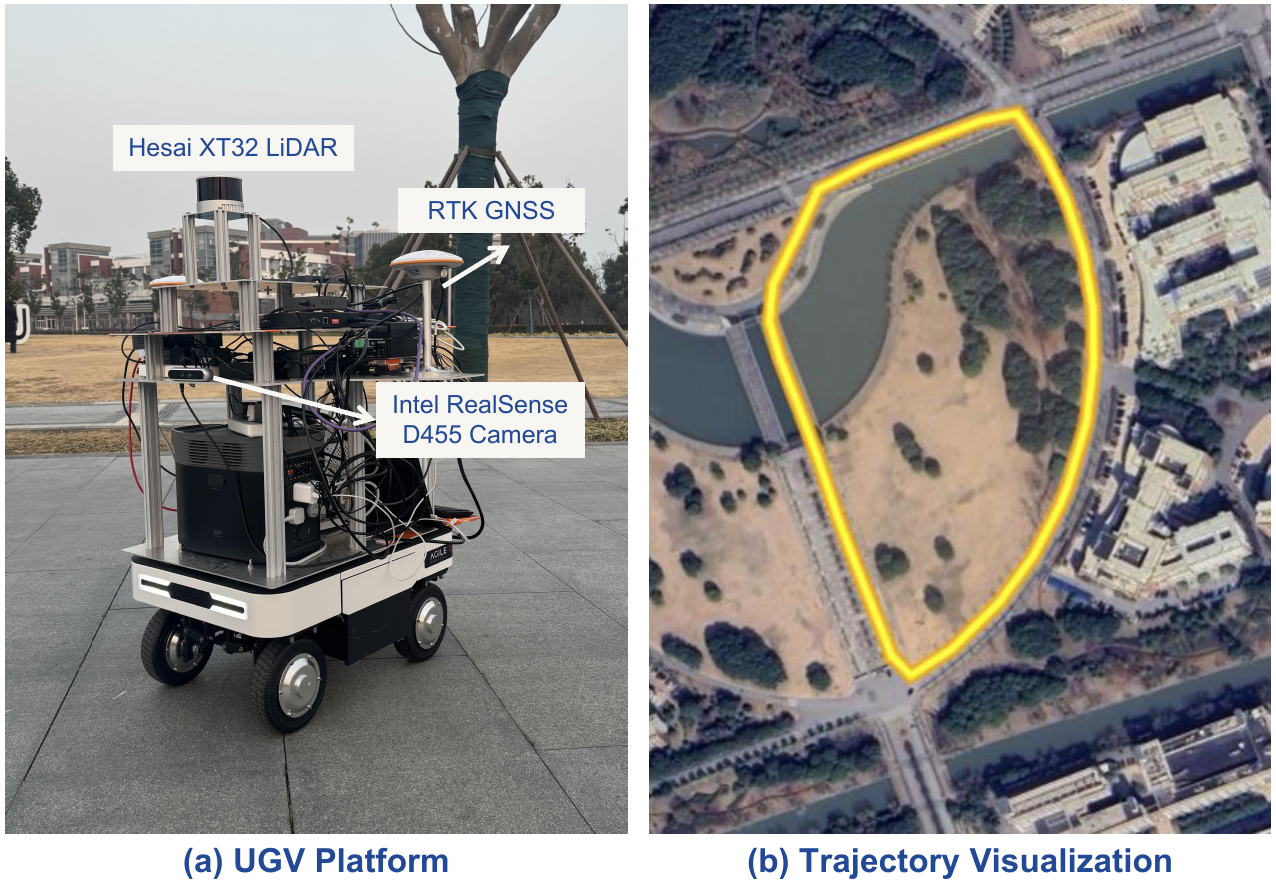}
  \caption{The unmanned ground vehicle (UGV) and collection trajectory for our self-collected data.}
  \label{fig:self-data}
  \vspace{-0.6cm}
\end{figure}

We train GRM with the lazy triplet loss commonly used in place recognition~\cite{qi2024gspr,xu2024explicit,zhou2023lcpr}. For each training sample, we construct a tuple consisting of an anchor query descriptor $\overline{\mathcal{D}}_{\text{query}}$, $N_\text{pos}$ positive descriptors $\overline{\mathcal{D}}_{\text{pos}}$, and $N_\text{neg}$ positive descriptors $\overline{\mathcal{D}}_{\text{neg}}$. The total training loss is formulated as:
\begin{align}
\begin{split}
    \mathcal{L} & = [N_\text{pos}(\beta+\max(d(\overline{\mathcal{D}}_\text{query}, \overline{\mathcal{D}}_\text{pos})))\\
    & -\sum_{N_\text{neg}}d(\overline{\mathcal{D}}_\text{query}, \overline{\mathcal{D}}_\text{neg})]_+,
\end{split}
\label{eq:loss}
\end{align}
where $[\cdot]_+$ denotes the hinge loss function, $\beta$ represents the constant margin, and $d(\cdot,\cdot)$ computes the Euclidean distance between descriptors. 

Upon training completion, we proceed to database construction, where global descriptors for all reference frames are computed and indexed in $\mathcal{M}$. 
During the following inference phase, the global descriptor of a query frame is generated to retrieve top-$k$ candidate matches from $\mathcal{M}$ via nearest neighbor search, resulting $\mathcal{C}$ for the following re-ranking operation.

\subsection{Re-Ranking Mechanism}
\label{sec:RRM}

Following the initial retrieval of the top-$k$ candidate matches $\mathcal{C}$, we introduce the training-free re-ranking mechanism (RRM) that exploits the strong point-tracking capability of VGGT. As illustrated in Fig.~\ref{fig:re-ranking}, our RRM comprises mask-guided keypoint extraction and confidence-aware correspondence scoring.
Specifically, we first utilize MobileSAM~\cite{zhang2023faster} to generate high-quality segmentation masks for the query image $\mathcal{I}_q$. We filter the regions with masks larger than the preset threshold, excluding indistinctive information such as sky and roads, while retaining semantically-rich regions as shown in Fig.~\ref{fig:re-ranking}(a). Within these remaining regions in $\mathcal{I}_q$, we apply the FAST detector to extract a set of robust keypoints $\mathcal{K}$.

Subsequently, we implement point tracking with VGGT and then calculate the correspondence score for each candidate in the GRM's top-$k$ matches $\mathcal{C}$, as shown in Fig.~\ref{fig:re-ranking}(b). 
Concretely, the query $\mathcal{I}_q$, a specific candidate $\mathcal{I}_j$ from $\mathcal{C}$, and the extracted keypoints $\mathcal{K}$ are collectively fed into the frozen VGGT followed by the DPT head. For each query point $\mathbf{y}_a \in \mathcal{K}$ in $\mathcal{I}_q$, the network predicts its corresponding point $\hat{\mathbf{y}}_a$ in $\mathcal{I}_j$, representing the identical 3D physical location across the two views of $\mathcal{I}_q$ and $\mathcal{I}_j$. 
Concurrently, a tracking confidence map $\mathcal{U}_j$ is automatically estimated by the DPT head to quantify the reliability of the correspondence. 
Then, our proposed tracking confidence aggregation (TCA) is implemented to calculate the correspondence score with three confidence-aware metrics. The first one is the median score $S_{\text{med}}$, which directly adopts the median value of the tracking confidence map to ensure robustness against outliers. The second one is the high-confidence ratio $S_{\text{high}}$, which measures the proportion of tracking points with confidence scores exceeding a predefined threshold. The last one is the consistency score $S_{\text{cons}}$, which evaluates the stability of the tracking by taking the inverse of the standard deviation of the confidence map, normalized via a \textit{tanh} function. These scores are specifically calculated and aggregated by:
\begin{align}
  S_{\text{med}} &= \text{median}(\mathcal{U}_j),  \\
  S_{\text{high}} &= \frac{1}{N_\text{kp}}\sum_{a=1}^{N_\text{kp}}\mathbb{I}(u_a > \tau),  \\
  S_{\text{cons}} &= \text{tanh}\left(\frac{\alpha}{\sigma_j + \epsilon}\right),\\
  S_{\text{total}}(\mathcal{I}_q, \mathcal{I}_j) &= \lambda_1 S_{\text{med}} + \lambda_2 S_{\text{high}} + \lambda_3 S_{\text{cons}}, 
  \label{eq:uncertrainty} 
\end{align}
where $N_\text{kp}$ denotes the number of keypoints $\mathcal{K}$, $u_{a}$ represents the confidence value for the $a$-th keypoint in $\mathcal{K}$ sampled from $\mathcal{U}_j$, and $\tau$ is the threshold to select high-confidence keypoints. $\mathbb{I}(\cdot)$ denotes the indicator function, which takes the value 1 if $u_a > \tau$ is satisfied and 0 otherwise.
$\sigma_j$ represents the standard deviation of $\mathcal{U}_j$, and $\alpha$ is a scalar hyperparameter for the consistency score. The total correspondence score $S_{\text{total}}(\mathcal{I}_q, \mathcal{I}_j)$ is the weighted sum of the aforementioned three metrics, where $\lambda_1$, $\lambda_2$, and $\lambda_3$ denote the corresponding balancing weights. As can be noted, TCA helps to justify the overall tracking quality between the query and candidates from GRM's global retrieval. This is because two observations in similar places hold better visual alignment, leading to more keypoint correspondences and higher tracking confidence than their counterparts in spatially distant locations. As shown in Fig.~\ref{fig:re-ranking}(b), candidate \textbf{B}, which is spatially closer to the query location, exhibits a significantly higher tracking correspondence score (0.92) compared to candidate \textbf{A} (0.10). With our proposed confidence-aware re-ranking, we effectively differentiate between true and false positives. 
Consequently, the top-$k$ candidates are re-ranked in descending order based on these scores for better place recognition performance.


\section{Experiments and Analyses}
\label{sec:experiments}

\subsection{Experimental Setups}
\label{sec:setups}

\subsubsection{Datasets}
We evaluate our method on three public outdoor autonomous driving datasets, including nuScenes~\cite{caesar2020nuscenes}, NCLT~\cite{carlevaris2016university}, and KITTI~\cite{geiger2013vision}. 
For the nuScenes dataset, the Boston-Seaport (BS) split is used for training and testing, while the Singapore-Onenorth (SON) and Singapore-Queenstown (SQ) splits are exploited for zero-shot evaluation. 
For the NCLT dataset, we select ``2012-02-04'' and ``2012-03-17'' sequences for training. During testing, we regard ``2012-01-08'' as the database, ``2012-06-15'', ``2013-02-23'', and ``2013-04-05'' as queries.
For KITTI, sequences 00, 02, and 05 are selected for loop-closure detection evaluation.
These dataset configurations follow the previous works~\cite{zhou2023lcpr, xu2024explicit, qi2024gspr, qi2025unimpr, lu2025ring, ma2023cvtnet}, offering reasonable data splits to evaluate long-term place recognition performance in large-scale autonomous driving environments.
In addition, we also report VGGT-MPR's performance on our self-collected data in real-world autonomous driving environments. As shown in Fig.~\ref{fig:self-data}, we utilized an unmanned ground vehicle equipped with a Hesai XT32 LiDAR, an Intel RealSense D455 camera, and an RTK GNSS to collect one query sequence (7,861 frames) and one reference sequence (7,847 frames). 

\begin{figure}[t]
  \centering
  \includegraphics[width=0.98\linewidth]{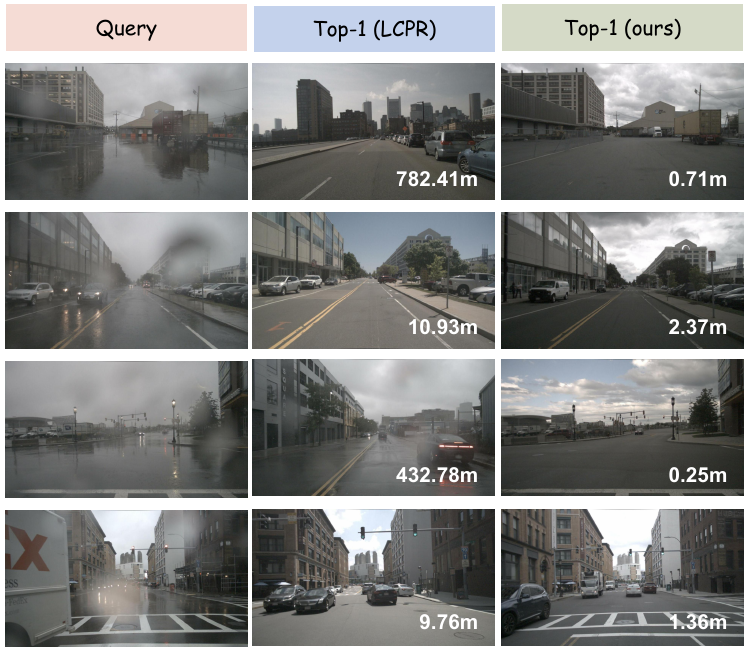}
  \caption{Visualization of retrieval results. The first column shows the current query image, the second and third columns show the top-$1$ places retrieved by LCPR and our method, respectively. The value in the bottom-right corner of each retrieved image indicates its distance to the query position.}
  \label{fig:results_compare}
  \vspace{-0.4cm}
\end{figure}

\begin{table*}[t]
\scriptsize
\setlength{\tabcolsep}{6.1pt}
\center
\captionsetup{aboveskip=2pt, belowskip=0pt}
\renewcommand\arraystretch{0.95}
\caption{Performance comparison on the NCLT dataset}
\begin{tabular}{l|c|cccc|cccc|cccc}
\toprule
\multicolumn{1}{l|}{\multirow{2}{*}{Approach}}  & \multicolumn{1}{l|}{\multirow{2}{*}{Modality}}   & \multicolumn{4}{c|}{\begin{tabular}[c]{@{}c@{}}Query: 2012-06-15\\ Database: 2012-01-08\end{tabular}} & \multicolumn{4}{c|}{\begin{tabular}[c]{@{}c@{}}Query: 2013-02-23\\ Database: 2012-01-08\end{tabular}}   & \multicolumn{4}{c}{\begin{tabular}[c]{@{}c@{}}Query: 2013-04-05\\ Database: 2012-01-08\end{tabular}}  \\ \cmidrule{3-14} 
\multicolumn{1}{c|}{} & \multicolumn{1}{c|}{} & AR@1  & AR@5  & AR@10  & AR@20  & AR@1  & AR@5  & AR@10  & AR@20  & AR@1  & AR@5  & AR@10  & AR@20\\ \cmidrule{1-14}
PatchNetVLAD & V  & 59.27 & 65.09 & 67.23 & 68.69 & 53.22 & 58.91 & 61.71 & 64.01 & 55.72 & 62.03 & 64.93 & 67.80 \\
CricaVPR & V  & 63.33 & 71.78 & 75.57 & 78.69 & 53.92 & 63.28 & 68.13 & 73.43 & 54.24 & 62.09 & 66.54 & 71.34 \\
CVTNet   & L  & \underline{84.14} & \underline{88.94} & \underline{90.92} & \underline{93.01} & \underline{84.90} & \underline{89.81} & \underline{91.77} & \underline{93.30} & \underline{82.25} & \underline{87.62} & \underline{89.67} & \underline{91.80} \\
OverlapMamba  & L  & 63.71 & 75.52 & 79.97 & 84.48 & 62.69 & 74.26 & 78.24 & 82.60 & 58.48 & 69.86 & 75.02 & 79.39 \\
\cmidrule{1-14}
MinkLoc++  & V+L  & 42.77 & 51.89 & 56.24 & 60.52 & 35.44 & 47.06 & 53.58 & 61.40 & 36.38 & 56.25 & 64.76 & 72.48 \\
LCPR   & V+L  & 68.63 & 76.84 & 80.23 & 83.63 & 68.37 & 76.15 & 79.77 & 83.12 & 60.58 & 69.25 & 73.07 & 77.45 \\
EINet  & V+L  & 71.52 & 81.07 & 84.70 & 87.75 & 57.09 & 68.11 & 73.35 & 78.20 & 53.48 & 64.14 & 68.62 & 72.95 \\
GSPR   & V+L  & 80.58 & 87.49 & 90.25 & 92.57 & 59.50 & 71.33 & 75.70 & 80.72 & 57.61 & 69.96 & 75.00 & 79.97 \\
\rowcolor{lightgray} VGGT-MPR (ours)  & V+L  & \textbf{86.29} & \textbf{93.03} & \textbf{94.83} & \textbf{97.53} & \textbf{86.06} & \textbf{90.73} & \textbf{92.62} & \textbf{95.38} & \textbf{82.44} & \textbf{89.15} & \textbf{90.41} & \textbf{94.06} \\ \bottomrule
\end{tabular}
\label{tab: comparison_nclt}
\\ \vspace{-0.15cm}
\vspace{-0.4cm}
\end{table*}

\subsubsection{Implementation Details}
We utilize the front-view camera image and LiDAR point clouds of each frame as the inputs of VGGT-MPR.
The dimension of the modality-specific descriptors is set to $D = 256$, resulting in a fused multimodal global descriptor $\overline{\mathcal{D}}\in\mathbb{R}^{512}$.
For all datasets, the distance thresholds for positive and negative samples are uniformly set to $9$\,m and $18$\,m, respectively.
For training, we employ the lazy triplet loss with $N_\text{pos}=2$ positives,  $N_\text{neg}=6$ negatives, and a margin $\beta=0.5$. During testing, we retrieve top-$30$ candidates in the database for each query. That is, we implement re-ranking for $k=30$ frames to refine recognition results. 
When calculating the correspondence score in RRM, we empirically set $\tau=0.7$ to select high-confidence keypoints, and $\alpha=0.1$ to scale the consistency score. Besides, we use $\lambda_1=0.45$, $\lambda_2=0.45$, and $\lambda_3=0.10$ to balance multiple metrics for the total correspondence score.
To quantify retrieval performance, we adopt the average recall (AR) metric, which is widely used in the literature. We report AR at different ranks (e.g., AR@1, AR@5, AR@10, and AR@20) on the test sets.
We train VGGT-MPR for 10 epochs, using the ADAM optimizer with an initial learning rate of 1e-5 for nuScenes, 5e-5 for NCLT and KITTI.
All of our experiments are conducted on a single NVIDIA A100 GPU.

\begin{table}[t]
\scriptsize
\setlength{\tabcolsep}{1.1pt}
\center
\captionsetup{aboveskip=2pt, belowskip=0pt}
\renewcommand\arraystretch{0.95}
\caption{Performance comparison on the KITTI dataset}
\begin{tabular}{l|ccc|ccc|ccc}
\toprule
\multicolumn{1}{l|}{\multirow{2}{*}{Approach}}  & \multicolumn{3}{c|}{00} & \multicolumn{3}{c|}{02}   & \multicolumn{3}{c}{05}  \\ \cmidrule{2-10} 
\multicolumn{1}{c|}{} & AR@1  & AR@5  & AR@10 & AR@1  & AR@5  & AR@10  & AR@1  & AR@5  & AR@10  \\ \cmidrule{1-10}
MinkLoc++   & 77.84 & 81.36 & 82.61 & 68.96 & 73.73 & 74.33 & 61.23 & 65.26 & 66.84 \\
LCPR    & 51.82 & 70.80 & 78.07 & 35.52 & 54.93 & 62.99 & 40.18 & 56.84 & 62.99 \\
EINet  & 70.15 & 78.04 & 78.64 & 59.61 & 61.02 & 69.37 & 53.49 & 59.96 & 64.05 \\
\rowcolor{lightgray} VGGT-MPR & \textbf{82.69} & \textbf{83.41} & \textbf{84.54} & \textbf{71.94} & \textbf{74.55} & \textbf{75.84} & \textbf{66.84} & \textbf{67.19} & \textbf{68.37} \\ \bottomrule
\end{tabular}
\label{tab:kitti}
\vspace{-0.3cm}
\end{table}

\subsection{Comparison With SOTA Methods}
\label{sec:performance}

\begin{table}[t]
\scriptsize
\setlength{\tabcolsep}{12.7pt}
\center
\captionsetup{aboveskip=2pt, belowskip=0pt}
\renewcommand\arraystretch{0.95}
\caption{Performance comparison on our self-collected data}
\begin{tabular}{l|cccc}
\toprule
Approach       & AR@1 & AR@5 & AR@10 & AR@20 \\ \cmidrule{1-5} 
MinkLoc++   & 62.67 & 72.45 & 77.28 & 80.26\\
LCPR    & 57.01 & 69.55 & 76.74 & 80.93\\
EINet  & 70.44 & 81.48 & 85.23 & 87.80\\
GSPR   & 47.88 & 65.31 & 74.76 & 78.35\\
\rowcolor{lightgray} VGGT-MPR & \textbf{76.05} & \textbf{87.52} & \textbf{92.59} & \textbf{94.17} \\ \bottomrule
\end{tabular}
\label{tab:self-collect}
\vspace{-0.3cm}
\end{table}

\begin{table}[t]
\scriptsize
\setlength{\tabcolsep}{10.2pt}
\center
\captionsetup{aboveskip=2pt, belowskip=0pt}
\renewcommand\arraystretch{0.95}
\caption{Ablation study on input modalities}
\begin{tabular}{cc|cccc}
\toprule
\multicolumn{1}{c}{\begin{tabular}[c]{@{}c@{}}Vision\\Input\end{tabular}} & \multicolumn{1}{c|}{\begin{tabular}[c]{@{}c@{}}LiDAR\\Input\end{tabular}} & AR@1 & AR@5 & AR@10 & AR@20 \\ \cmidrule{1-6} 
\ding{55}   & \ding{51}  & 89.12 & 98.84 & 99.32 & 99.32 \\ 
\ding{51}  & \ding{55}   & 94.01 & 99.10 & 99.54 & 99.77 \\  
\rowcolor{lightgray} \ding{51}  & \ding{51}  & \textbf{98.28} & \textbf{99.76} & \textbf{99.93} & \textbf{99.93} \\  
 \bottomrule
\end{tabular}
\label{tab:modality}
\vspace{-0.3cm}
\end{table}

\begin{table}[t]
\scriptsize
\setlength{\tabcolsep}{6.0pt}
\center
\captionsetup{aboveskip=2pt, belowskip=0pt}
\renewcommand\arraystretch{0.95}
\caption{Ablation study on VGGT functions in GRM}
\begin{tabular}{cc|cccc}
\toprule
\multicolumn{1}{c}{\begin{tabular}[c]{@{}c@{}}Depth\\Densification\end{tabular}} & \multicolumn{1}{c|}{\begin{tabular}[c]{@{}c@{}}Visual Embedding\\Extraction\end{tabular}} & AR@1 & AR@5 & AR@10 & AR@20 \\ \cmidrule{1-6} 
\ding{55}   & \ding{55}  & 92.51 & 98.32 & 99.42 & 99.73 \\ 
\ding{51}   & \ding{55}  & 95.84 & 99.10 & 99.56 & 99.79 \\ 
\ding{55}   & \ding{51}  & 97.91 & 99.68 & 99.89 & 99.90 \\ 
\rowcolor{lightgray} \ding{51}  & \ding{51}   & \textbf{98.28} & \textbf{99.76} & \textbf{99.93} & \textbf{99.93}  \\  
 \bottomrule
\end{tabular}
\label{tab:VGGT_ablation}
\vspace{-0.3cm}
\end{table}

\begin{table}[t]
\scriptsize
\setlength{\tabcolsep}{9.5pt}
\center
\captionsetup{aboveskip=2pt, belowskip=0pt}
\renewcommand\arraystretch{1.1}
\caption{Ablation study on the re-ranking mechanism}
\begin{tabular}{l|cccc}
\toprule
Approach       & AR@1 & AR@5 & AR@10 & AR@20 \\ \cmidrule{1-5} 
w/o RRM (nuScenes)     & 97.21 & 99.42 & 99.78 & 99.84 \\  
\rowcolor{lightgray}w/ RRM (nuScenes)   & \textbf{98.28} & \textbf{99.76} & \textbf{99.93} & \textbf{99.93}  \\  \midrule
w/o RRM (NCLT)     & 85.22 & 91.60 & 94.34 & 96.92 \\  
\rowcolor{lightgray}w/ RRM (NCLT)   &\textbf{86.29} & \textbf{93.03} & \textbf{94.83} & \textbf{97.53} \\  \midrule
w/o RRM (KITTI)     & 81.93 & 83.18 & 83.98 & 84.89 \\  
\rowcolor{lightgray}w/ RRM (KITTI)   & \textbf{82.69} & \textbf{83.41} & \textbf{84.54} & \textbf{85.45} \\ 
 \bottomrule
\end{tabular}
\label{tab:rerank_ablation}
\vspace{-0.4cm}
\end{table}

\begin{figure}[t]
  \centering
  \includegraphics[width=0.98\linewidth]{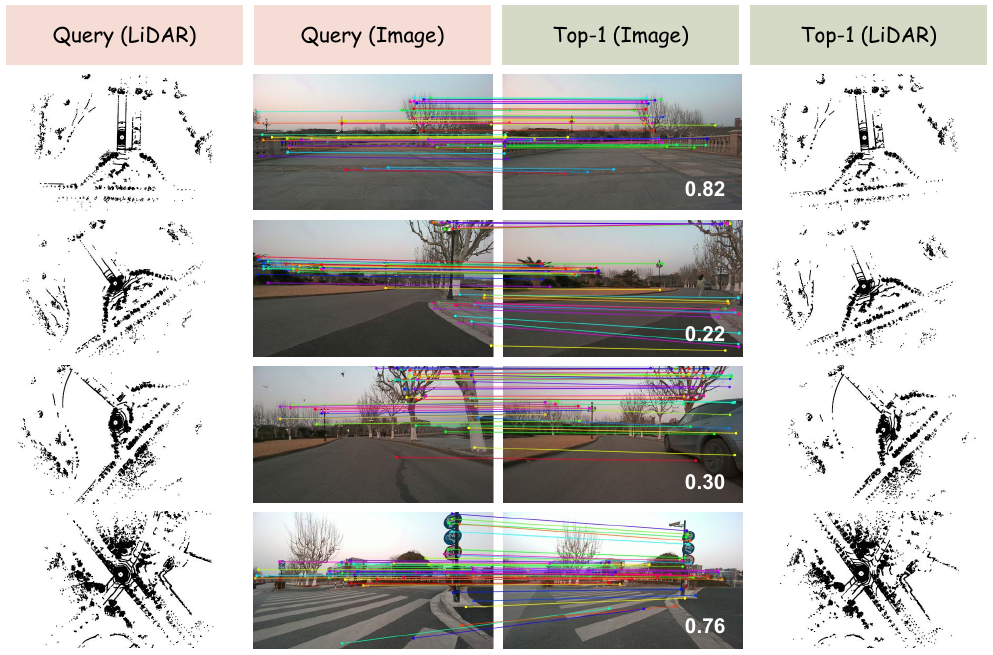}
  \caption{Visualization of retrieval results on the self-collected dataset. The first two columns present the query LiDAR point clouds and camera image, while the last two columns display the corresponding retrieved matches.}
  \label{fig:results_self-data}
  \vspace{-0.3cm}
\end{figure}

\subsubsection{Evaluation on Public Datasets}

We compare VGGT-MPR with a wide spectrum of representative SOTA baselines, categorized into vision-based (PatchNetVLAD~\cite{hausler2021patch}, CricaVPR~\cite{lu2024cricavpr}), LiDAR-based (CVTNet~\cite{ma2023cvtnet}, OverlapMamba~\cite{luo2025overlapmamba}), and multimodal approaches (MinkLoc++~\cite{komorowski2021minkloc++}, LCPR~\cite{zhou2023lcpr}, EINet~\cite{xu2024explicit}, GSPR~\cite{qi2024gspr}).

As shown in Tab.~\ref{tab: comparison}, VGGT-MPR demonstrates superior place recognition performance across all the test splits of the nuScenes dataset. Specifically, it outperforms the secondary baseline, GSPR, by 7.96\% (AR@1), 0.88\% (AR@5), and 0.35\% (AR@10) on the BS split.
In addition, VGGT-MPR shows better generalization to unseen autonomous driving scenarios when tested on the SON and SQ splits in a zero-shot manner. 
Furthermore, Fig.~\ref{fig:results_compare} compares our proposed VGGT-MPR with LCPR, a baseline that also claims occlusion and changing lighting conditions. As can be observed, in challenging scenarios involving weather changes, viewpoint variations, and significant occlusion, our method can retrieve better places in the database for the given query compared to LCPR.
This basically benefits from the foundation model capabilities of the integrated VGGT, compared to all the other unimodal and multimodal baselines whose parameters are optimized from scratch. We further present the generalization performance of baseline methods and our approach on the NCLT dataset. 
Concretely, we trained models only on sequences ``2012-02-04'' and ``2012-03-17'', and test them on ``2012-06-15'', ``2013-02-23'', and ``2013-04-05'' to evaluate their performance under limited training data in long time spans. 
As shown in Tab.~\ref{tab: comparison_nclt}, our VGGT-MPR holds better recognition performance across all test sequences. It can still achieve AR@1 of 86.06\% and 82.44\% on the reference sequences ``2013-02-23'' and ``2013-04-05'' over one year after the query sequence, which are significantly higher than the secondary baseline CVTNet.

Tab.~\ref{tab:kitti} further compares the performance of MPR baselines and our approach on KITTI, all using only a purely vision branch. The experimental results indicate that in scenarios where LiDAR sensor data is unavailable, our method is still robust enough to effectively retrieve the most correct places. GSPR~\cite{qi2024gspr} is not evaluated here as it does not have a vision branch separated from its multimodal fusion strategy.

\subsubsection{Evaluation on Our Self-Collected Data}

Moreover, we report VGGT-MPR's place recognition performance on our self-collected data. Concretely, we directly transfer the model pretrained on nuScenes to implement zero-shot inference. As shown in Tab.~\ref{tab:self-collect}, our proposed VGGT-MPR achieves the best recognition performance compared to the SOTA MPR baselines, validating its good generalization ability. 
Fig.~\ref{fig:results_self-data} presents the correspondence scoring results of four example retrievals, which further demonstrates that our re-ranking mechanism generates reasonable scores for similar places.

\subsection{Ablation Studies}
\label{sec:ablation}

\begin{figure}[t]
  \centering
  \includegraphics[width=0.98\linewidth]{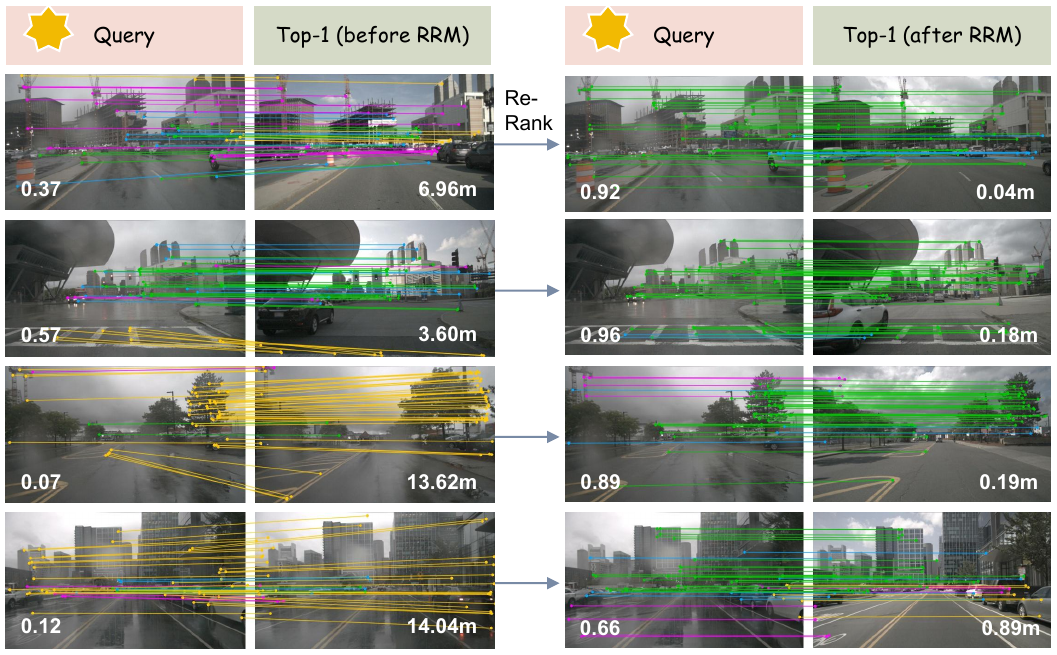}
  \caption{Qualitative comparison of top-$1$ retrieval results before and after re-ranking. Colored lines indicate keypoint tracking correspondences between the query and the candidate image, where the color denotes the tracking confidence from VGGT, i.e., high confidence ($>0.75$) in green, medium-high confidence ($0.50$--$0.75$) in yellow, medium-low confidence ($0.25$--$0.50$) in purple, and low confidence ($<0.25$) in blue. The correspondence score and real distance are displayed in the bottom-left and bottom-right corners of each query-candidate pair.}
  \label{fig:results_reranking}
  \vspace{-0.4cm}
\end{figure}

\begin{table}[t]
\scriptsize
\setlength{\tabcolsep}{13.8pt}
\center
\captionsetup{aboveskip=2pt, belowskip=0pt}
\renewcommand\arraystretch{0.95}
\caption{Ablation study on the number of retrieved candidates}
\begin{tabular}{c|cccc}
\toprule
Number       & AR@1 & AR@5 & AR@10 & AR@20 \\ \cmidrule{1-5} 
1     & 97.21 & - & - & - \\ 
5     & 97.76 & 99.44 & - & - \\  
10    & 97.96 & 99.69 & 99.70 & - \\  
15    & 97.73 & \textbf{99.76} & 99.82 & - \\  
20    & 97.78 & 99.60 & 99.86 & 99.88 \\  
25    & 97.80 & 99.65 & 99.86 & 99.88 \\  
\rowcolor{lightgray}30   & \textbf{98.28} & \textbf{99.76} & \textbf{99.93} & \textbf{99.93}  \\  
 \bottomrule
\end{tabular}
\label{tab:candidate_ablation}
\vspace{-0.3cm}
\end{table}

\subsubsection{Input Modalities}
Tab.~\ref{tab:modality} presents comparative results of different input modalities, including vision-only, LiDAR-only, and the full multimodal VGGT-MPR configuration on the BS split of the nuScenes dataset. As can be seen, the fusion of both modalities results in the highest recognition accuracy, demonstrating the coordinated benefits of multimodal information. Notably, the vision branch exhibits a more significant contribution than the LiDAR branch, which can be attributed to the plausible VGGT-based visual embedding extraction.

\subsubsection{VGGT Functions in GRM}
We further ablate the effectiveness of VGGT adapted in our GRM's submodules on nuScenes. In Tab.~\ref{tab:VGGT_ablation}, the variant without any VGGT functions directly uses 2D stacked convolutions to extract features from camera images and LiDAR range images, which holds the worst recognition performance. Either VGGT-based depth densification or visual embedding extraction leads to performance enhancement, while the improvement from the latter function is more prominent. This finding is consistent with the results in the ablation study in Tab.~\ref{tab:modality}, where the VGGT's foundation model capability of providing strong geometrically-rich priors substantiates the dominant contribution of the vision modality.

\subsubsection{Re-Ranking Mechanism}
Tab.~\ref{tab:rerank_ablation} also highlights the performance gains attributed to the proposed RRM with different datasets, including the BS split of nuScenes, sequence ``2012-06-15'' of NCLT, and sequence 00 of KITTI. As can be seen, incorporating the re-ranking operation results in consistent performance improvement, which demonstrates that the point tracking of VGGT effectively extracts spatial correspondence between the query and retrieved candidates, leading to the refinement of the final recognition results with our devised confidence-aware scoring.
Fig.~\ref{fig:results_reranking} further presents qualitative top-$1$ retrieval results before and after applying the proposed re-ranking mechanism. Although the initially retrieved candidates often exhibit a similar global appearance to the query, their local correspondences are frequently noisy and inconsistent, as reflected by a large number of low-confidence tracking links. In contrast, after re-ranking, the selected top-$1$ candidate demonstrates more coherent and stable tracking patterns, characterized by a higher proportion of high-confidence correspondences. This leads to significantly increased aggregated correspondence scores. That is, our proposed TCA helps to effectively distinguish visually similar but geographically distinct locations based on tracking reliability.

\subsubsection{Number of Retrieved Candidates for Re-Ranking}

Moreover, we explore the performance change with the increasing number of retrieved candidates $k$ for re-ranking. Note that we can not report AR@$x$ when $x>k$, since there are not enough candidates to evaluate.
Tab.~\ref{tab:candidate_ablation} shows that the performance gain from re-ranking becomes more pronounced as the number of retrieved candidates increases on nuScenes. This is because an expanded retrieval scope increases the likelihood of encompassing the ground truth, thereby enabling the re-ranking mechanism to identify more accurate candidates.


\section{Conclusion}

In this paper, we presented VGGT-MPR, a novel framework that reinterprets the VGGT as a unified geometric engine for robust multimodal place recognition. By fusing visual perception with 3D structural awareness, our global retrieval module effectively enhances visual embeddings and densifies sparse LiDAR data to capture complementary modality strengths. Moreover, we introduced a training-free re-ranking mechanism that exploits explicit cross-view point tracking to ensure spatial consistency, effectively refining retrieval results without additional parameter optimization. Extensive evaluations on large-scale benchmarks demonstrate that VGGT-MPR significantly outperforms state-of-the-art methods, exhibiting solid generalization ability in zero-shot scenarios and robustness against severe environmental variations. This work highlights the significant potential of adopting visual foundation models to advance multimodal place recognition in autonomous driving.


\bibliographystyle{ieeetr}
\footnotesize{\bibliography{new}}

\end{document}